\documentclass[journal]{IEEEtran}
\usepackage{graphicx}
\usepackage{graphics} 
\usepackage{epsfig} 
\usepackage{mathptmx} 
\usepackage{times} 
\usepackage{amsmath} 
\usepackage{amssymb}  
\usepackage{caption}
\usepackage{subcaption}
\usepackage{float}
\restylefloat{table}
\usepackage[tableposition=top]{caption}
\restylefloat{table}

\usepackage{url}
\urldef{\mailsa}\path|{alfred.hofmann, ursula.barth, ingrid.haas, frank.holzwarth,|
\urldef{\mailsb}\path|anna.kramer, leonie.kunz, christine.reiss, nicole.sator,|
\urldef{\mailsc}\path|erika.siebert-cole, peter.strasser, lncs}@springer.com|

\newcommand{\RNum}[1]{\uppercase\expandafter{\romannumeral #1\relax}}

\ifCLASSINFOpdf
\else
 
\fi

\begin{document}
%
\title{An Unsupervised  Method for Detection and Validation of The Optic Disc and The Fovea}

\author{Mrinal~Haloi,  Samarendra~Dandapat, and~Rohit~Sinha
\thanks{M.Haloi, S. Dandapat, R. Sinha is with the Department
of Electrical and Communication Engineering, Indian Institute of Technology, Guwahati,India
 e-mail: (h.mrinal, samaren,rsinha)@iitg.ernet.in. / mrinal.haloi11@gmail.com}
}


\maketitle


\begin{abstract}
In this work, we have presented a novel method for detection of retinal image features, the optic disc and the fovea, from colour fundus photographs of dilated eyes for Computer-aided Diagnosis(CAD) system. A saliency map based method was used to detect the optic disc followed by an unsupervised probabilistic Latent Semantic Analysis for detection validation. The validation concept is based on distinct vessels structures in the optic disc. By using the clinical information of standard location of the fovea with respect to the optic disc, the macula region is estimated. Accuracy of 100\% detection is achieved for the optic disc and the macula on MESSIDOR and DIARETDB1 and 98.8\% detection accuracy on STARE dataset. 

\end{abstract}


\section{Introduction}
Analysis of retinal image for detection of its pathological \cite{c23} and non-pathological features is very important for automatic computer aided detection and diagnosis of retinal diseases. With the emergence of medical image analysis research for faster and accurate analysis by reducing cost and time, researchers developing computer software to facilitate easier medical treatment. Use of computer aided diagnosis will help doctors in remote areas and faster analysis of retinal images, lower cost of treatment by reducing manpower. Most of the works on retinal image analysis uses retinal images obtained from fundus photography \cite{c20, micro} by dilating pupil. The optic disc, the fovea, blood vessels and veins are main features of retinal image. Different types of retinal diseases effect those features. Age related macular degeneration causes defect of macula region also by diabetic retinopathy. Hard and soft exudates are pathological features responsible for diabetic retinopathy. Damage of optic disc is result of Glaucoma diseases [19], one major cause of vision loss. Optic disc cup and neuro retinal rim surface areas ratio, determine presence and progression of Glaucoma. Diameter changes of retinal arteries and veins are associated with different cardiovascular diseases. Thinning of the arteries and widening of the veins, result in increased risk of stroke and myocardial infraction \cite{aao}.

Detection of the optic disc, the fovea, blood vessels and veins are very important for pathological analysis of retinal image \cite{c23}. Since severity of diseases depends on location of features of corresponding diseases with respect to those essential features. Also some pathological features occurs in specific areas. The optic disc (OD) can be observed as bright part of normal retinal image, and the fovea as the darkest region. In pathological images sometimes it’s very hard to detect the optic disc and the macula region due to abnormalities caused by different diseases. In most of the recent works on detection accuracy of the optic disc and the macula region in pathological images is very low, no proper validation of detection region is presented. Eventhough the location of the macula can be estimated from the optic disc location but due to age related macular degeneration that region may be severely effected, to find whether the region is normal or that of pathological eyes, proper validation is needed.
 For efficient detection of the OD and the fovea consideration of adverse illumination variation and damaging for these features due to eye diseases need to be addressed. Also illuminace variation linked with imaging setup. Fig. 1 shows a typical retinal image with the optic disc, the macula region, the blood vessels and pathological feature exudates.\\
\begin{figure}
  \centering
      \includegraphics[width=3.5in,height=2.3in]{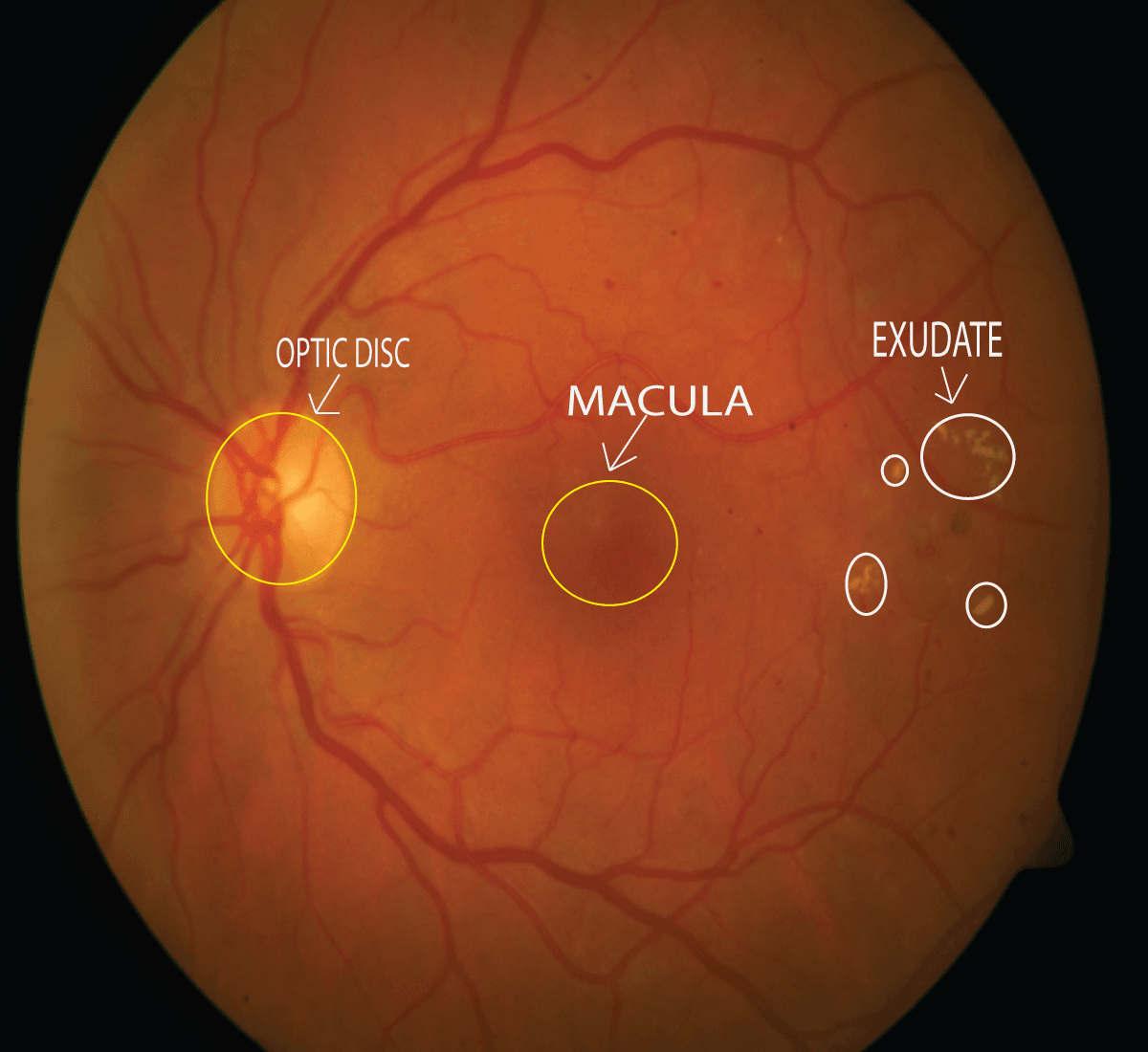}
\caption{Retinal Features}
\end{figure}

Retinal image analysis is a mature area, lots of work has been done in this area. But still a complete state of art system for performing all analysis with high accuracy is not achieved. People have used image processing and machine learning based approach for detection and classification of various features. It is not possible to give a complete analysis of these works. A few works were selected for discussing.

Lu et al. \cite{lu} have used line operator to detect circular brightness structure of the optic disc. Their method failed to address large illuminance variation effect with neighbouring regions and limited to clear circular brightness structure.
Matched filter based method was proposed by Abdel et al. \cite{abdel}, they first preprocesses the images by illumination normalization and histogram equalization. Their algorithm performance depends on the vessel segmentation. And will be effected by adverse illuminance variation.
Using retinal vessel direction informtion Foracchia et al. \cite{forc} proposed a geometrical parametric model for the optic disc detection. The algorithm need vessel segmented images and hence vessel segmentation performance effect its results. 

A probability map based localization of the optic disc is presented by Budai et al. \cite{budai}, they have constructed two probability map. One is brightness based from the brightness of image and other is vessel segmentation based. And combination of these two map was used to locate the optic disc. And this method failed to address the problems of illuminance variation and difficulty involves in pathological images.
A model based approach was used by Li et al. \cite{li} to detect the optic disc, the macula and exudates. Principal component based method was used to locate the optic disc and active shape based method for shape detection of the optic disc and the fovea. 

In this work, two new novel method for detection of retinal features is developed.  The optic disc detection is based on saliency map that can capture significant variation of local structure. Once the potential probable location of the optic disc detected, a probabilistic Latent Semantic Analysis based unsupervised method was used to validate whether the region is the optic disc or not, this is based on specific vasculature structure in that region. This method gives us 100 \% accuracy in optic disc detection in different challenging images with pathological symptoms. This algorithm is described in section \RNum{2}(A) and \RNum{2}(B).  The macula region detection by using the information of the optic disc location and the main courses of blood vessels is described in section \RNum{2}(C). 

\section{Method}
This method comprises of mainly two parts. In the first stage detection and validation of the optic disc is performed, while the fovea detection is dependent on OD detection. A complete overview of the method used in this work is explained in the Fig. 2. 

\begin{figure}
  \centering
      \includegraphics[width=3.3in,height=2.8in]{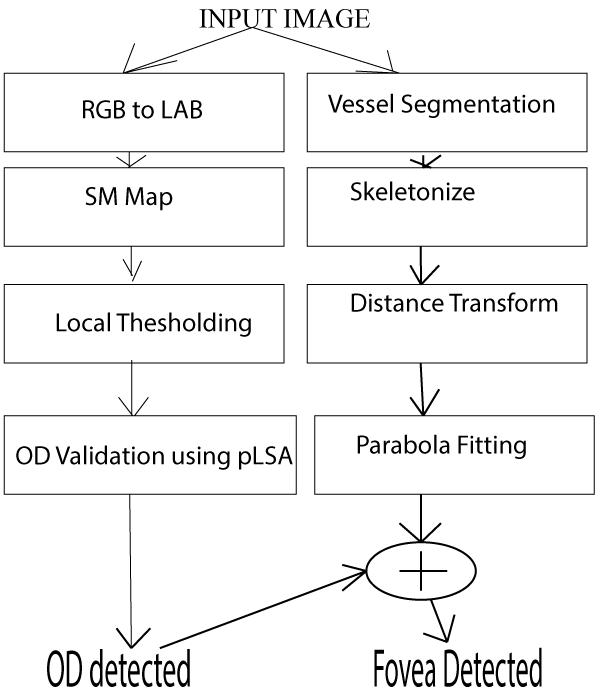}
\caption{Method Overview}
\end{figure}

\subsection{Optic Disc Detection}
The optic disc is one of the most important anatomical structures of retina. The optic disc is also known as the blind spot, because there are no light sensitive rods or cones to respond to a light stimulus. The retinal arteries and veins emerge from the left of the optic disc.
Its central white depression called the physiologic cup and horizontal diameter of it should not exceed 1/2 that of entire disc, otherwise it’s a sign of pathologic optic disc cupping reasons behind glaucoma.
For detection of optic disc a saliency region detection algorithm \cite{smap} was used to identify salient region of the image, based on image local structure variation. For this first image will be convert to CIE Lab colour space, where luminance value from $L$ channel of image and colour value from $a$ and $b$ channel can be seperated.
Conversion process is described \cite{lab} as follows.

\begin{equation}
\left[ \begin{array}{c} X \\ Y\\Z \end{array} \right] = \begin{bmatrix} 0.4887180 & 0.3106803 & 0.2006017 \\  0.1762044  & 0.8129847 & 0.0108109 \\  0.0000000 & 0.0102048 & 0.9897952 \end{bmatrix} \times \left[ \begin{array}{c} R \\ G \\B \end{array} \right]
\end{equation}

\begin{equation}
\begin{aligned}
L  = 116[ h( \frac{Y}{Y_W} )] - 16 \\
a = 500[h( \frac{X}{X_W} ) - h(\frac{Y}{Y_W})] \\
b = 200[h( \frac{Y}{Y_W} ) - h(\frac{Z}{Z_W})] 
\end{aligned}
\end{equation}

\begin{equation}
 h(q)= 
\begin{cases}
    \sqrt[3]{q} ,& \text{if } q > 0.008856\\
    7.787q + 16/116,              & \text{otherwise}
\end{cases}
\end{equation}

Because of optic disc colour and structuring variation from neighbourhood variation, saliency map (SM) can efficiently detect it. Saliency is computed on the basis of variation of local contrast of a image patch with respect to neighbourhood. Even if the colour vaiation is low but with strong structural variation SM can capture OD. This process is repeated at different scales for getting better accuracy in constructing the map and keeping finer details. The contrast based saliency of a given pixel at position $(i,j)$ is $c_{i,j}$ and computed as follows \cite{smap}.
\begin{equation}
c_{i,j} =  D[(\frac{1}{N1}\displaystyle\sum\limits_{p=1}^{N1} v_{p}), (\frac{1}{N2}\displaystyle\sum\limits_{q=1}^{N2} v_{q})]
\end{equation}

Where D is Euclidean distance between the average vectors of two non-overlapping patch $P_{1}$ and $P_{2}$ and  with total number of pixels as $N1$ and $N2$. If these two vectors are correlated, Mohalanobis distance will be used. Also, $v_{p} and v_{q}$ are vector of features of two pixels corresponding to two different regions. Vector corresponding to each pixels have three features, Luminance and two colour 'a' and 'b' from CIE Lab colour space channel.
\begin{equation}
v_{i} = [L_{i},a_{i},b_{i}]
\end{equation}
Size of patch $P_{1}$ is taken as 9x9 and that of $P_{2}$ is calculated from the equation. Input image have c number of columns.
\begin{equation}
\frac{c}{2} \geq (N_{P_{2}}) \geq \frac{c}{8} 
\end{equation}

Final saliecny map values is computed as sum of saliency map of that image at different scales,
\begin{equation}
smap_{i,j} = \displaystyle\sum\limits_{S} c_{i,j}
\end{equation}
\begin{figure}
  \centering
      \includegraphics[width=3.3in,height=2.3in]{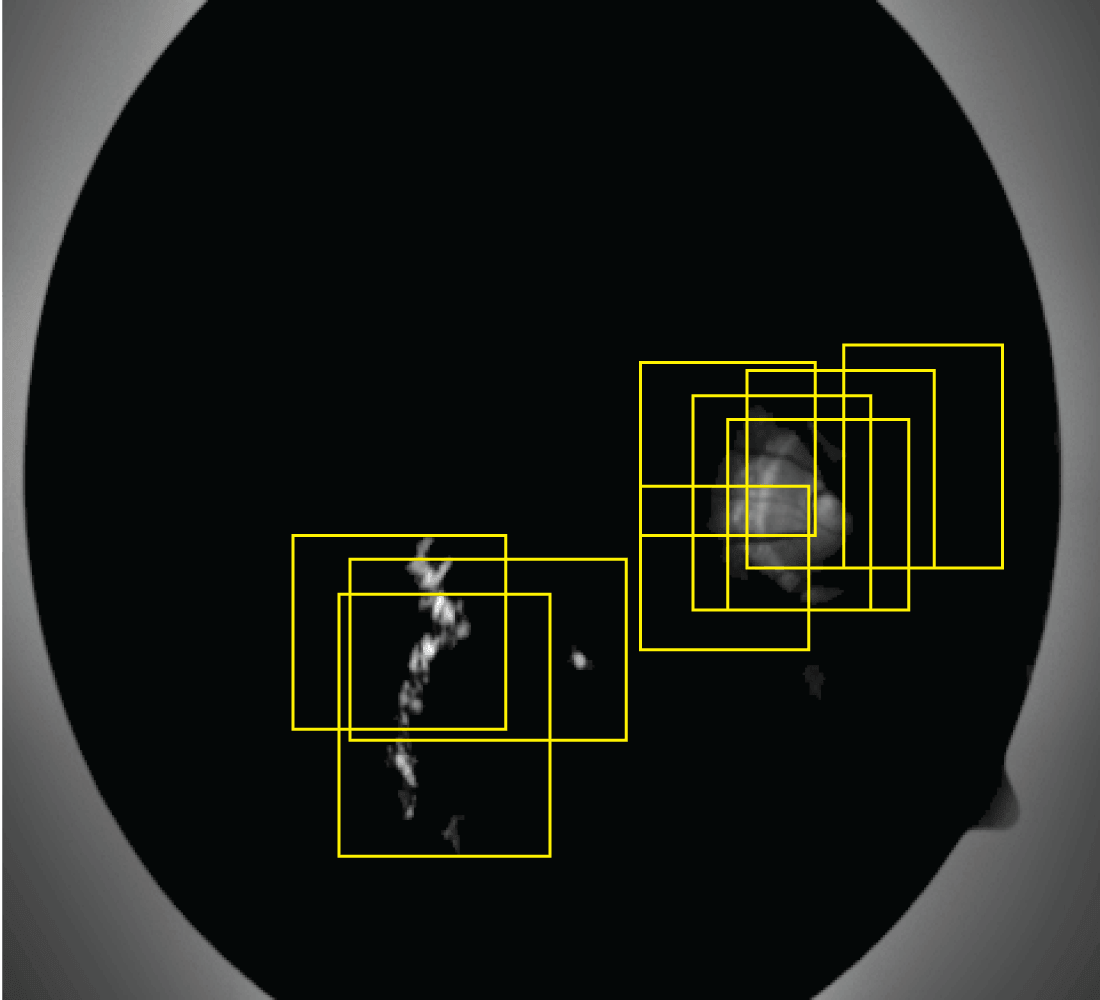}
\caption{Window selection from Segmented Region}
\end{figure}
For segmentation of relevant region from saliency map, patch based segmentation techniques was used by using the relation as follows. For each patch, mean $m_{p}$ and standard deviation $\sigma_{p}$ is computed. If $SM_{i,j}$ from equation (8) is greater than 1 then the pixels will be included in final interest map, otherwise it will be discarded.
\begin{equation}
SM_{i,j} = \frac{smap(i,j) - m_{p}}{\sigma_{p}}
\end{equation}
In the final map $SM$, due to presence of different pathological symptoms in retinal images, other features along with optic disc also got detected. To get final location of optic disc a validation method will be applied.

\subsection{Validation of detected Optic Disc}
Final saliency map may have different regions, optic disc or non-optic disc. To validate whether a region is optic disc we have used a unsupervised Probabilistic latent Semantic Analysis classification algorithm. The optic disc structures consist of a complex pattern of vessels originate from it, no other part of retina has this structures, this concept have been exploited for classification and this method is independent of luminance of optic disc region. Even if due to some pathological region luminance is depreciated our method still can accurately detect the optic disc. The specific vasculature structure may be defected due to several pathological problems, to address this problem, the part based classification model was exploited. Our model comprised six classes, five classes for the optic disc and its parts and the sixth class is other retinal features for differentiation. The optic disc is divided into four parts as shown in Fig. 4. This formulation efficiently detect the optic disc even if its structure is defected.
\begin{figure}
  \centering
      \includegraphics[width=3.3in,height=2.6in]{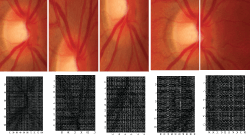}
\caption{OD part and corresponding HOG features}
\end{figure}
In the testing phase around each regions multiples window shifted to right, left, top and bottom was chosen. Variable window sizes as shown in Fig. 3 and window pixels value will be that of original image at those pixels location. Below stages involves in designing the classifier have been discussed. 
 Since image is a very high dimensional data, pre-process it to reduce its dimensionality by using visual codebook formation method. Each image will be represented as a bag of visual words. By using histogram of words concept, each image will be converted to a document with previously designed vocabulary.

\subsubsection{Feature Extraction}
For extracting meaningful edges information from images HOG \cite{hog} descriptor was used.
For object recognition HOG is a very popular local descriptor. This descriptor capture fine structure of images and suitable for object recognition. HOG computation based on gradient magnitude and phase.  Window with $16 \times 16 $ is selected with 50 \% overlap with neighbouring window, then it is further divided into $2 \times 2$ cells having size $8 \times 8$. For each window gradient phase is quantized into equally spaced 9 bins and gradient magnitude was used to determine values of each bin. 
 \begin{figure*}

 \center

  \includegraphics[width=4.5in, height = 2.4in]{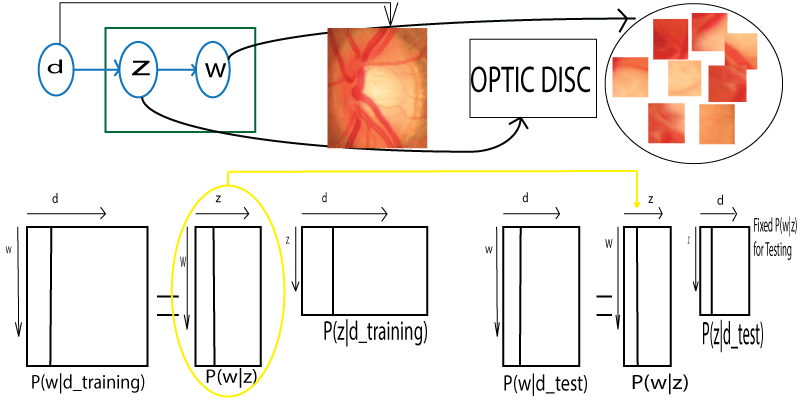}

  \caption{PLSA algorithm idea}

  \label{AAA}

\end{figure*}

\subsubsection{LLC Codebook Formation}
For the formation of visual words locality constrained linear coding \cite{llc} based method was used. This algorithm generate similar codes for similar descriptors by sharing bases. Locality also leads to sparsity. Idea based locality importance more than sparsity is used and given by below optimization problem. Here X is a D dimensional local descriptors ,$X = [x_{1},x_{2},...,x_{N}] \in \Re^{DxN}$ and $B = [b_{1},b_{2},...,b_{M}] \in \Re^{DXM}$ is $M$ dimensional codebook. Process is describes as follows.  
\begin{equation}
min_{c}\Sigma_{i=1,N}||x_{i} - Bc_{i}||^{2} + \lambda||d_{i} \odot c_{i}||^{2}
\end{equation}
\begin{equation}
s.t. 1^{T}c_{i} = 1, \forall i
\end{equation}
\begin{equation}
d_{i} = exp(\frac{dist(x_{i},B)}{\sigma})
\end{equation}
Where $\odot $ denotes the element-wise multiplication, and $d_{i} \in \Re^{M}$ is the locality adaptor that gives different freedom for each basis vector proportional to its similarity to the input descriptor$ x_{i}$. Also $dist(x_{i},B)$ is the Euclidean distance between $x_{i}$  and B. Shift invariance nature is confirmed by the constraint equation (3). In our case we will have 4608 dimension HOG descriptors $x_{i}$for each image. And a codebook $B$ of size 113 words was formed. Each image was expressed as a combination of these words. Each image was represented as histogram of visual words. Fig. 6 Depicts codebook formation scenario.
\begin{figure}
  \centering
      \includegraphics[width=3.3in,height=2.6in]{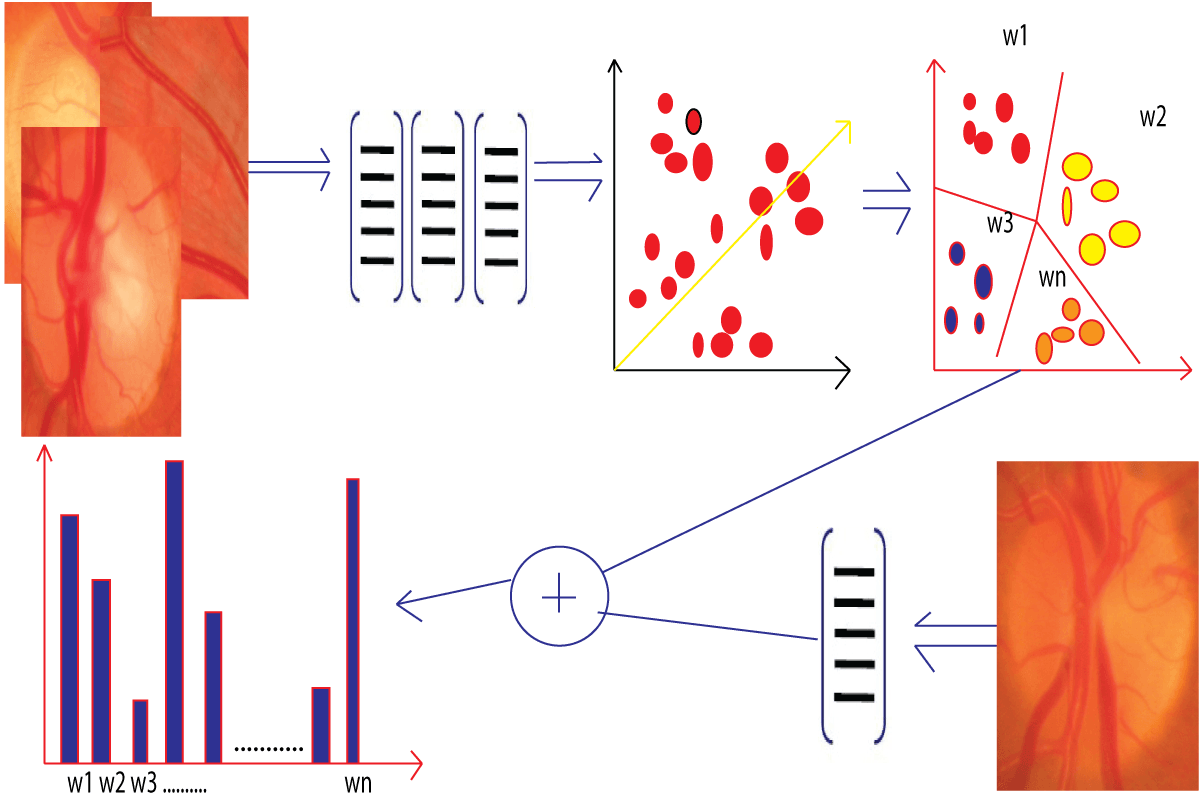}
\caption{Codeword formation}
\end{figure}
   
\subsubsection{PLSA model}
Probabilistic latent semantic analysis \cite{plsa} is a topic discovery model, its concept based on latent variable analysis. An image also can be considered as a collection of topics. Every image can be considered as a text document with words from a specific vocabulary. The vocabulary will be formed by using LLC algorithm on HOG features from training image. Suppose a collection of N images(document) $D ={d_{1},d_{2},...,d_{N}}$ is avilable, and corresponding vocabulary with size $N1$ is $W={w_{1},w_{2},...,w_{N1}}$. And let there be $N2$ topic $Z={z_{1},z_{2},...,z_{N2}}$ Model parameter are computed using expectation maximization method. In Fig. 5 Concept of pLSA model and its training and testing process is shown.

\begin{equation}
P(z|d,w) = \frac{P(z)P(d|z)P(w|z)}{\Sigma_{z'} P(z')P(d|z')P(w|z')}
\end{equation}
\begin{equation}
P(w|z) = \frac{\Sigma_{d}n(d,w)P(z|d,w)}{\Sigma_{d,w'}n(d,w')P(z|d,w')}\\
\end{equation}
\begin{equation}
P(d|z) = \frac{\Sigma_{w}n(d,w)P(z|d,w)}{\Sigma_{d',w}n(d',w)P(z|d',w)}\\
\end{equation}
\begin{equation}
P(z) = \frac{\Sigma_{d,w}n(d,w)P(z|d,w)}{R}\\
\end{equation}
\begin{equation}
R \equiv\Sigma_{d,w}n(d,w)
\end{equation}

\subsubsection{Fuzzy-KNN classification}
Since our framework based on topic modelling and discovery, fuzzy KNN classification \cite{fuzzy} techniques was used for getting appropriate topic of a test image corresponding to training images. Fuzzy kNN perform better than traditional KNN algorithm, because it depends on weight of neighbours.
In the training stage we have calculated $P(w|z)$, which was used as input for testing algorithm to compute $P(z|d_{test})$.
After that a K- nearest neighbour algorithm was used to classify these image by using probability distribution $P(z|d_{train})$.
\begin{equation}
u_{i}(x) = \frac{\displaystyle\sum\limits_{j=1,K}u_{ij}(\frac{1}{||x - x_{j}||^{\frac{2}{(m-1)}}})}{\displaystyle\sum\limits_{j=1,K}(\frac{1}{||x - x_{j}||^{\frac{2}{(m-1)}}})}
\end{equation}
$u_{i}$ is value for membership strength to be computed, and $u_{ij}$ is previously labelled value of i th class for j th vector.
Final class label of the query point x is computed as follows:
\begin{equation}
u_{0}(x) = arg max_{i}(u_{i}(x) )
\end{equation}

\subsection{Macula Localization}
The macula is the central region of the retina situated at the posterior pole of the eye, between the superior and inferior temporal arteries. Its centre at a distance of 2.5 D from optic disc centre. D is optic disc diameter. Fovea is located at the centre of macula and this part responsible for specialized high acuity vision. Fovea region is exclusively made up of cones, without it fine details could not be seen. Age-related macular degeneration, diabetic macular edema etc. are most common disorder of macula region. 
 As per study and observation fovea the centre part of macula is located at a distance 2.5 D along the axis of symmetry of a parabola which vertex is at the centre of optic disc. Finally a parabola was fitted to the main courses of blood vessels, process is depicted in Fig. 7. 

\subsubsection{Vessel Point Selection and Curve Fitting}
For extracting blood vessels from image, a algorithm described in this work \cite{vesselmap} was used. In the first stage the binary map of vessels was extracted. And then connected component analysis was used to remove small unconnected parts as a pre-processing step. To get centreline of blood vessels by removing border a morphological skeletonize algorithm \cite{skel} based algorithm was used. Also we have computed distance transform \cite{dist} of binary vessels map. Final threshold for getting rid of small vessels and vein is computed using cumulative histogram count and real data volume. Since the real  interest was extract the main courses blood vessels, a point wise multiplication of skeleton and distance map was used to get potential candidate points.

A parabolic model is fitted to the main course of blood vessels using least squares non linear optimization algorithm.
\begin{equation}
y = \frac{x^2}{4r sin\theta}
\end{equation}
r and $\theta$ are two constants, which are calculated using Newton non linear least squares optimization method as described follows. Suppose ( $x_{i},y_{i}$ ) are vessels coordinates, 

\begin{equation}
minimize \displaystyle\sum\limits_{i=1,m} (y_{i} - f(x_{i}))^2 
\end{equation}

\begin{figure}
  \centering
      \includegraphics[width=3.4in,height=1.9in]{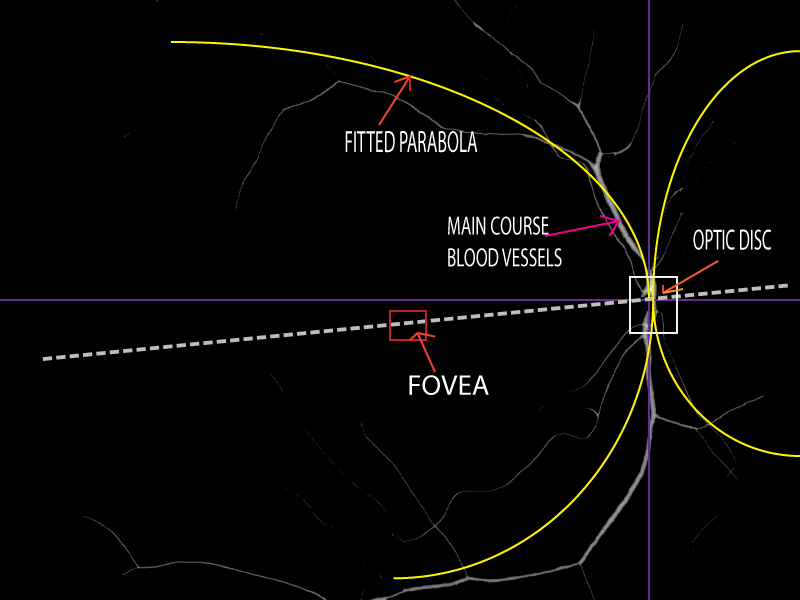}
\caption{Parabolic model Fitting}
\end{figure}

Once the position of the fovea is estimated we need to estimate whether the macula region is infected to any diseases. For this a template matching method have been applied and healthy eyes macula region was used as standard template. A window of size $1.5D \times 1.5D$ centred on the fovea proved to be efficient for this task and computed its distance (error) from standard template. If error is very high then the macula region is infected with age related macular degeneration.

\begin{table}[H]
  \centering
  \begin{tabular}{*{20}{c}}
\hline
Database & \# Pathological Img & Resolution & Acc(\%)\\
\hline
Messidor  & 300 & $1000 \times 1504$ & 100\\  
\hline
DIARETDB1 & 89 & $1152 \times 1500 $ & 100\\  
\hline
STARE & 81 & $605 \times 700$ & 98.8\\
\hline
\end{tabular} 
  \caption{Result of the optic disc and the Macula}
\end{table}

\begin{table}[H]
  \centering
  \begin{tabular}{*{20}{c}}
\hline
Methods & \# Images  & Acc(\%) & \# Failed img\\
\hline
Prposed Method & 81& 98.8 & 1\\
\hline
Abdel et al. \cite{abdel} & 81 & 98.8  & 1\\
\hline
Foracchia et al. \cite{forc} & 81 & 97.5 & 2 \\
\hline
Lu et al. \cite{lu} & 81 & 96.3 & 3\\
\hline
\end{tabular} 
  \caption{Result Comparisons of Different Optic Disc detection method on STARE dataset}
\end{table}

\section{RESULT and DISCUSSIONS}
For analysing the accuracy of this method on different set of images, publicly available MESSIDOR \cite{messi}, STARE[ \cite{stare}and DIARETDB1 \cite{diaret} dataset was experimented. Images with various difficulty from these databses for was testes to evaluate this algorithm accuracy. These datasets include image with pathological features such as haemorrhages, exudates and microaneurysms. These datasets are provided by expert in Ophthalmology with proper annotation of features location in images. These three datasets includes 400 retinal images with variety of pathological symptoms and captures under different conditions. Also camera used to capture those images for all these databases are different. We have used MATLAB platform in a windows 8.1 machine with intel i7 processor to test this method.
 
In Fig. 3, result of saliency map is shown with possible selection windows, it is clear that due to pathological symptoms it also capture irrelevant features. To get exact location of the optic disc a validation process in the detected areas was performed with multiple windows of size [122, 112], around the detected region. 

\begin{figure}
  \centering
      \includegraphics[width=3.1in,height=1.3in]{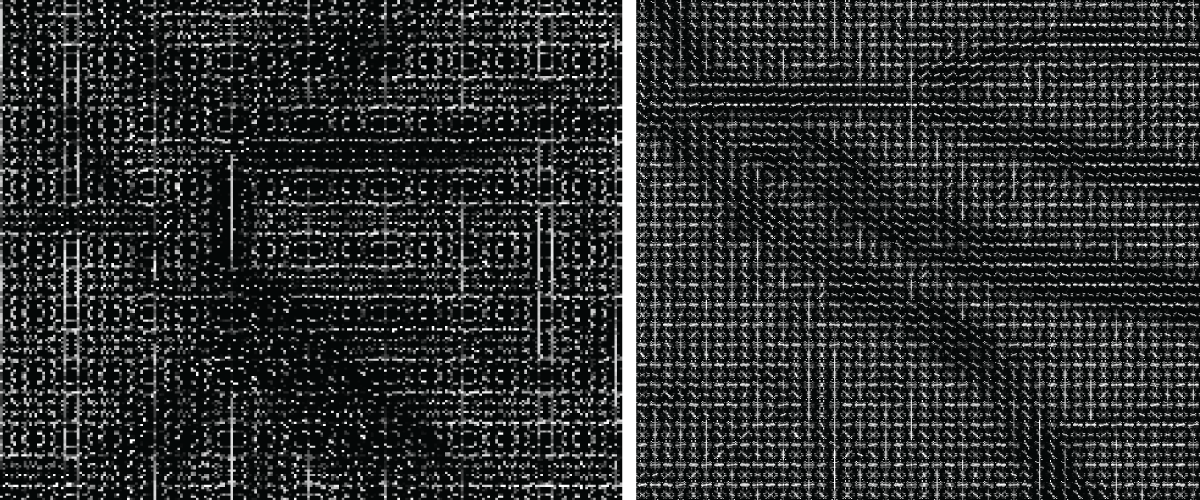}
\caption{Left: Optic disc Hog feature, Right: non optic disc element}
\end{figure}

\begin{figure*}
        \centering
        \begin{subfigure}[b]{0.33\textwidth}
                \includegraphics[width=1.9in,height=1.2in]{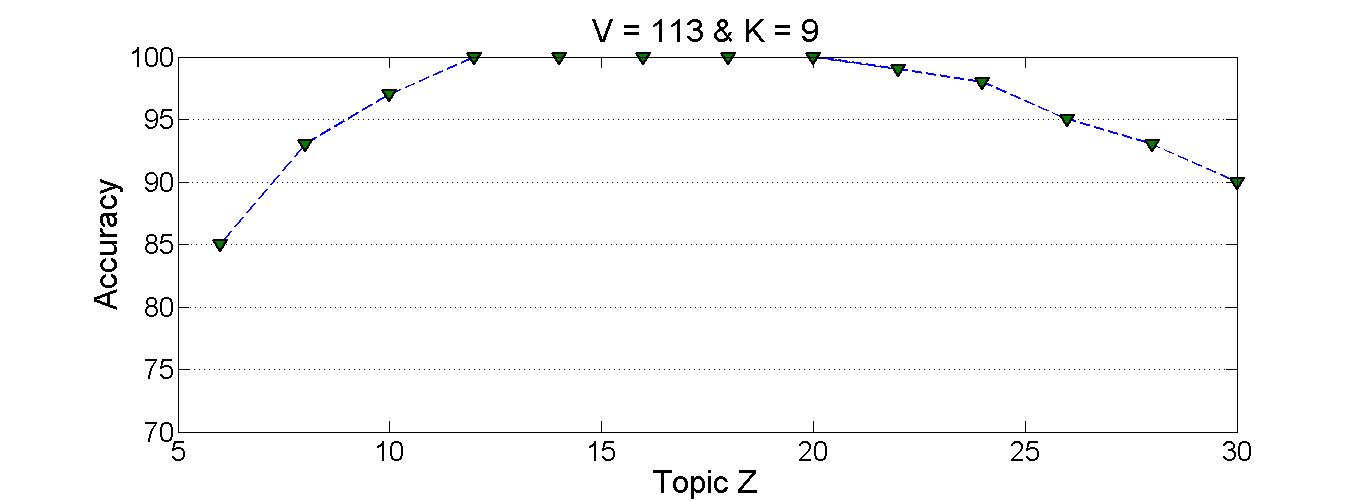}
                \caption{(a) Accuracy vs Z}
                \label{fig:cnnn}
        \end{subfigure}
        \begin{subfigure}[b]{0.33\textwidth}
                \includegraphics[width=1.9in,height=1.2in]{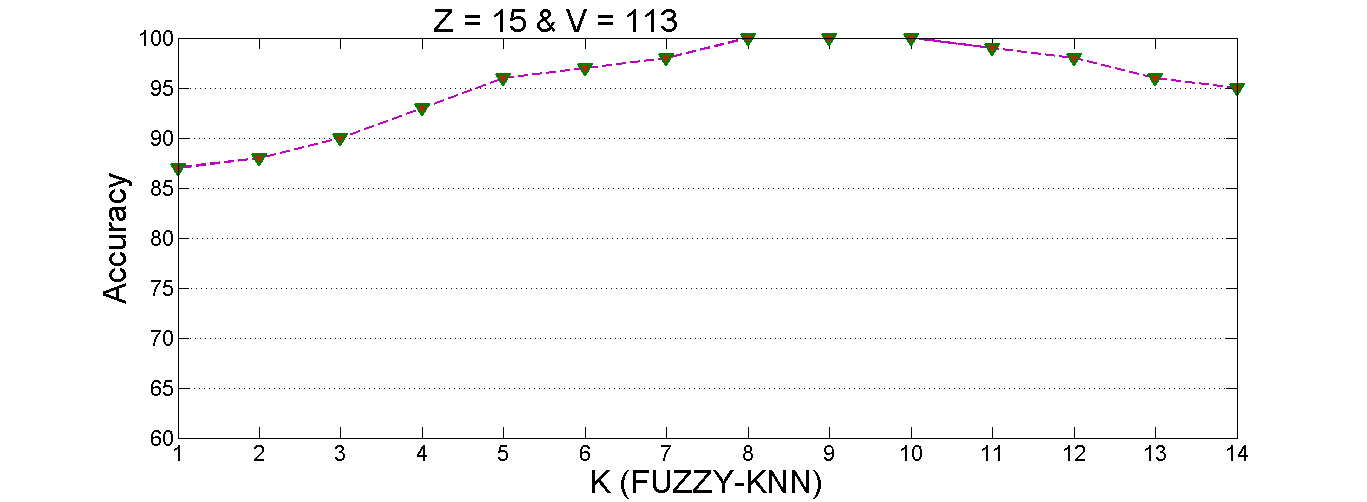}
                \caption{(b) Accuracy vs K}
                \label{fig:mcnn}
        \end{subfigure}
        \begin{subfigure}[b]{0.30\textwidth}
                \includegraphics[width=1.9in,height=1.2in]{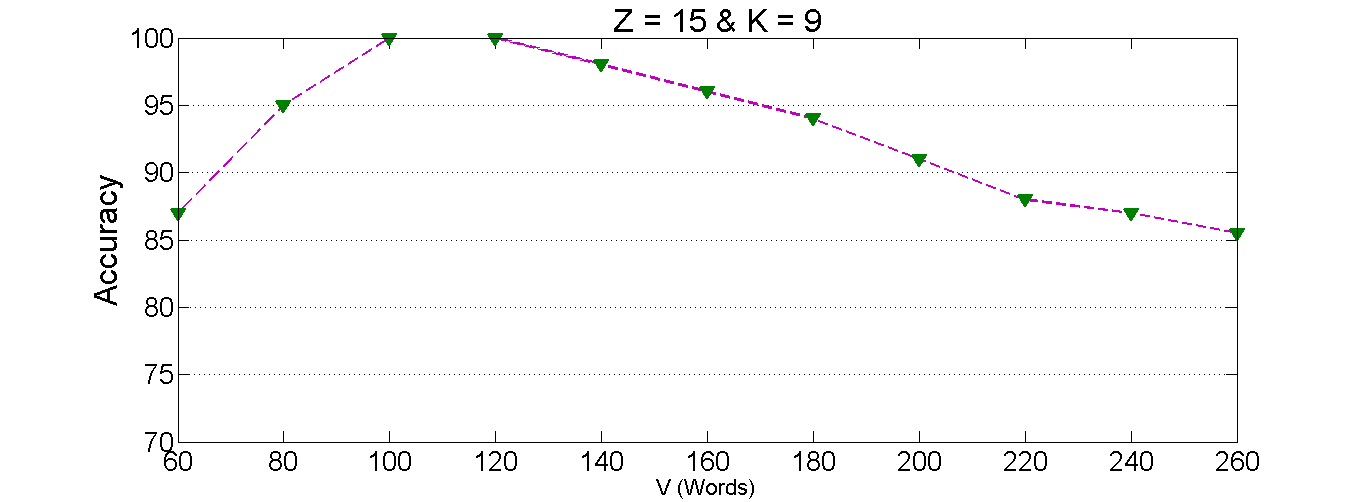}
                \caption{(c) Accuracy vs V}
                \label{fig:plsa}
        \end{subfigure}
        \caption{Parameters variation of pLSA}\label{imagess}
\end{figure*}

\begin{figure}
  \centering
      \includegraphics[width=3.4in,height=1.3in]{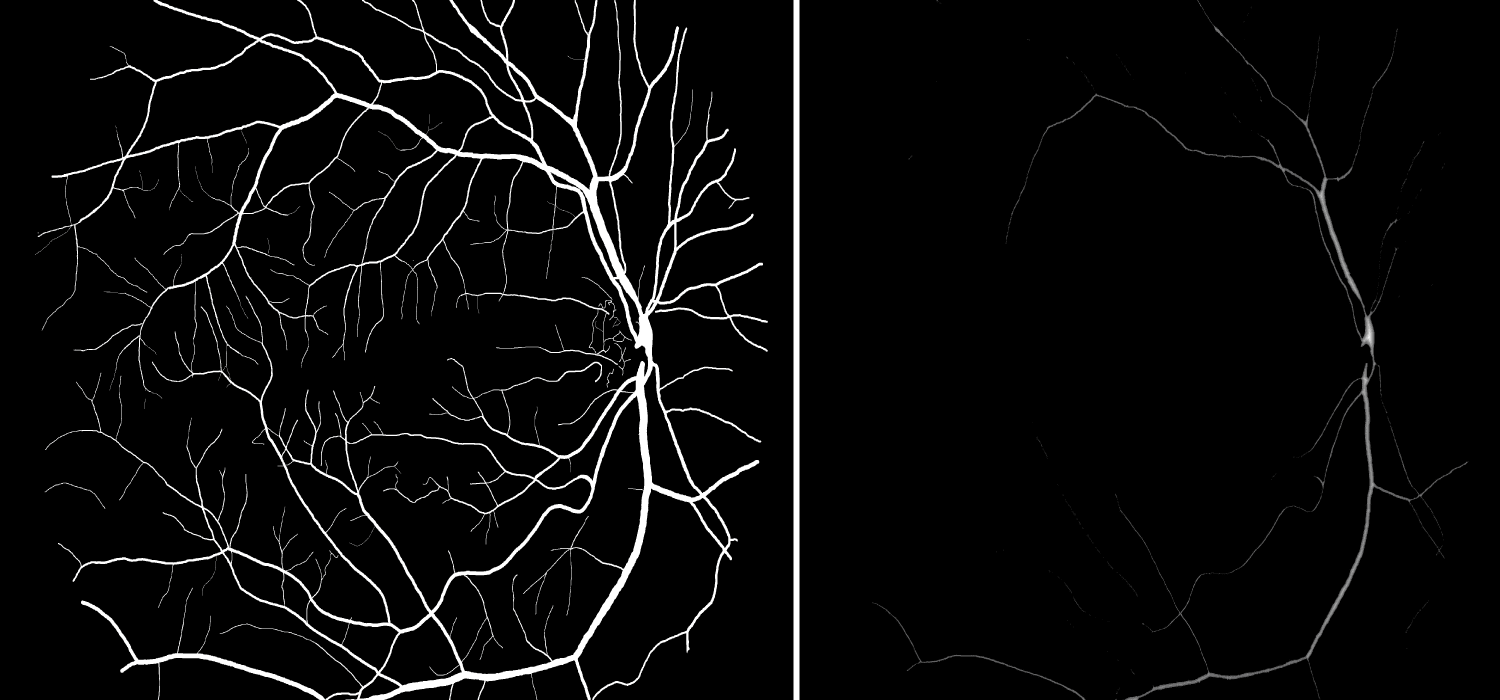}
\caption{Main courses of vessel after processing}
\end{figure}
\begin{figure}
  \centering
      \includegraphics[width=3.4in,height=2.4in]{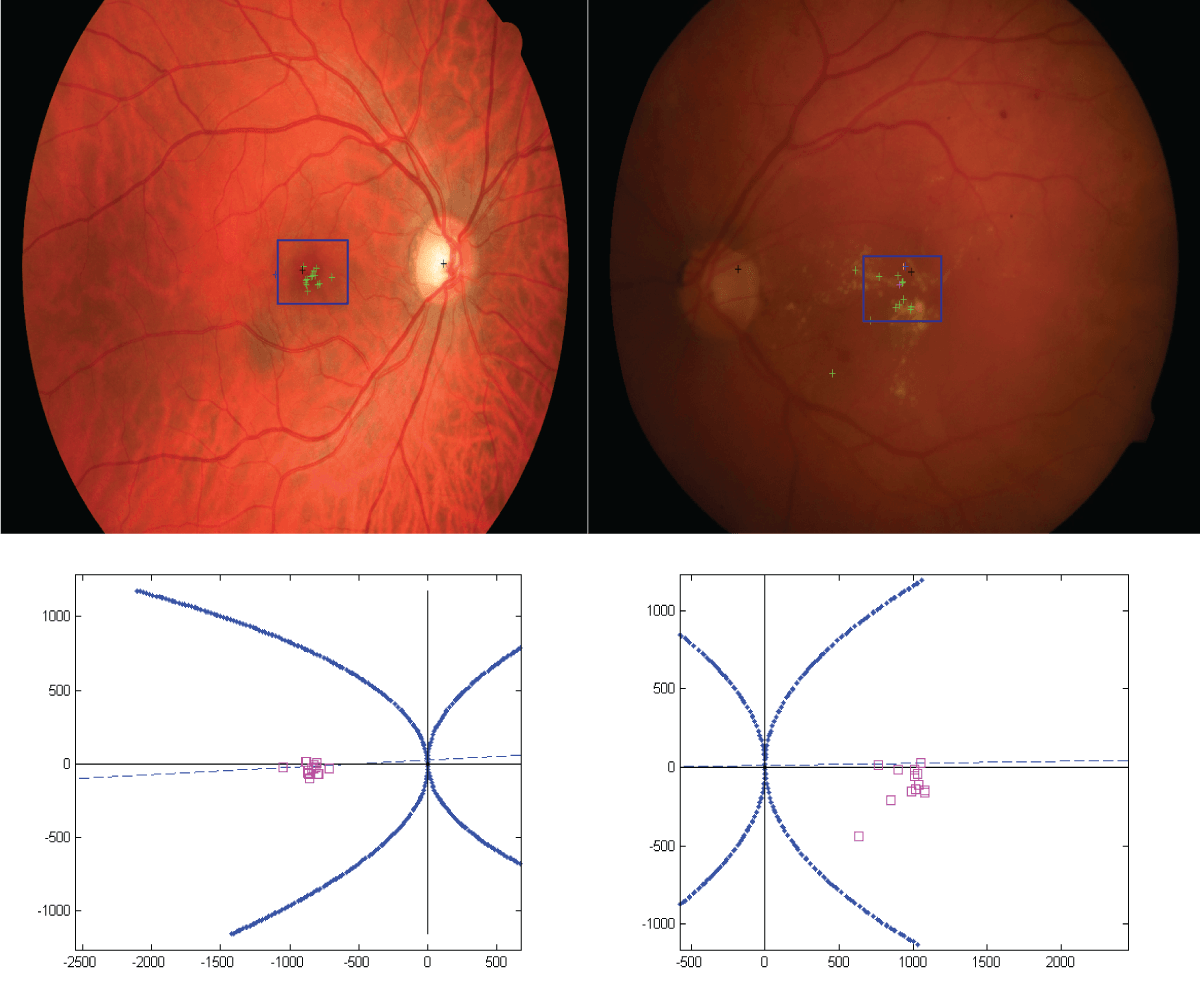}
\caption{Detected macula and fitted parabola}
\end{figure}

\begin{figure*}
 \center

  \includegraphics[width=6.5in, height = 2.4in]{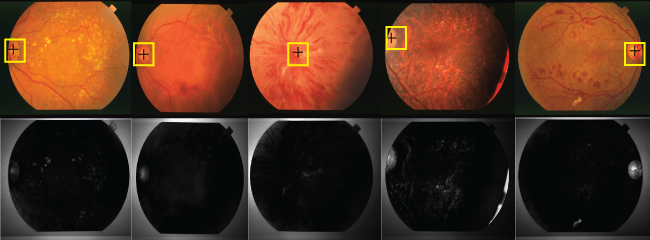}

  \caption{Optic disc detection result on STARE dataset, Top row: Image with detected OD, Bottom row: Corresponding Saliency Map}

  \label{AAA}

\end{figure*}

\begin{figure}
  \centering
      \includegraphics[width=3.4in,height=2.6in]{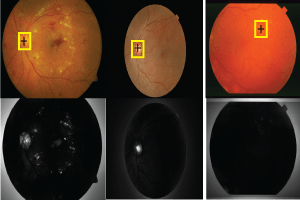}
\caption{Detected Optic Disc on DIARETDB1V2}
\end{figure}

For validation of the optic disc detection, the PLSA classifier is trained with 400 training images, these images are selected to include all type of possible difficulties. From each images the optic disc and its four parts is selected for training, and each image is resized to [122,112] using bilinear interpolation. From each window HOG feature as shown in Fig. 4 and Fig. 8 is extracted. Use of part based model improved the accuracy of this method and can detect OD even if it is half visible in the image. HOG feature can effectively capture structure of vessels. And non-optic disc windows are selected as negative for training the complete classifier. Classifier accuracy varies on the number of topics and words used in PLSA model, in this work classifier is trained using 15 topics, gives optimal accuracy.

Also value of number of nearest neighbours (K) in fuzzy-KNN effect accuracy. In this work optimal accuracy is obtained using 15 topics and with a visual dictionary of size 113. Optimal value of K used in this work is 9. Accuracy variation with respect to pLSA and fuzzy-KNN parameters is shown in Fig. 9. Fig . 9 (a) shows accuracy variation with respect to number of topics keeping number of words and number of nearest neighbours value fixed. Similarly Fig. 9(b) and Fig. 9 (c) shows variation with respect to K and V while keeping respective parameters fixed as shown.

 Detection of optic disc gets 100 \% accuracy over Diaretdb1v2 and Messidor databases. Our algorithm performs well in much degraded images with low light and severe pathological symptoms. In Fig. 12, detection result of the optic disc on pathological images (STARE dataset) with extreme illuminance and structure variation is shown. Also Fig. 12 left two columns shows some result in DIARETDB1 dataset. Table. \RNum{1} Gives detection accuracy of selected test images to include various difficulty from different dataset. A comparions with recent methods is shown in Table \RNum{2} on STARE dataset. This method perform very well with accuracy of 98.8\% on STARE.

In Fig. 10 Detected blood vessels from a image is shown and its main course of blood vessels obtained from morphological skeletonize and thresholding operation. Since main course of blood vessels have higher thickness and length than any other vessels and vein it becomes easier to extract them using as mentioned algorithm. Once the main course of blood vessels is extracted it become easy to fit parabola and locate the fovea.  
Result of macula location detection and curve fitting in normal and pathological images is shown in Fig. 11. It can be seen from the figure, that for pathological images with age related macular degeneration, macula's position can be detected and requires validation to confirm about defect. The process of confirming pathology in the macula is very important from the perspective of possible eye vision loss. For macular detection we have also 100 \% accuracy in normal images, this accuracy depends on the optic disc and main course of blood vessels locations. The fovea, which is centre of the macula region is located at a distance of 2.5 D of the optic disc diameter D.

\subsection{Luminance and Contrast Invariance}
An extensive evaluation is performed in images with varying luminosity and contrast. It has been observed that final Saliency map (SM) is robast to change of luminosity and contrast. In addition to that HOG feature values doesn't change with global contrast and luminosity variation. Fig. 14 clearly depicts the luminosity and contrast invariance, first coulmn shows the original image with color distribution and SM. First row and rest of the coloumn images were obtained by varying luminosity and contrast of original image at different lavel and second row shows their respective color disribution. It can be seen that final SM capture all features irrespective of variances. 
\begin{figure}
  \centering
      \includegraphics[width=3.5in,height=3.0in]{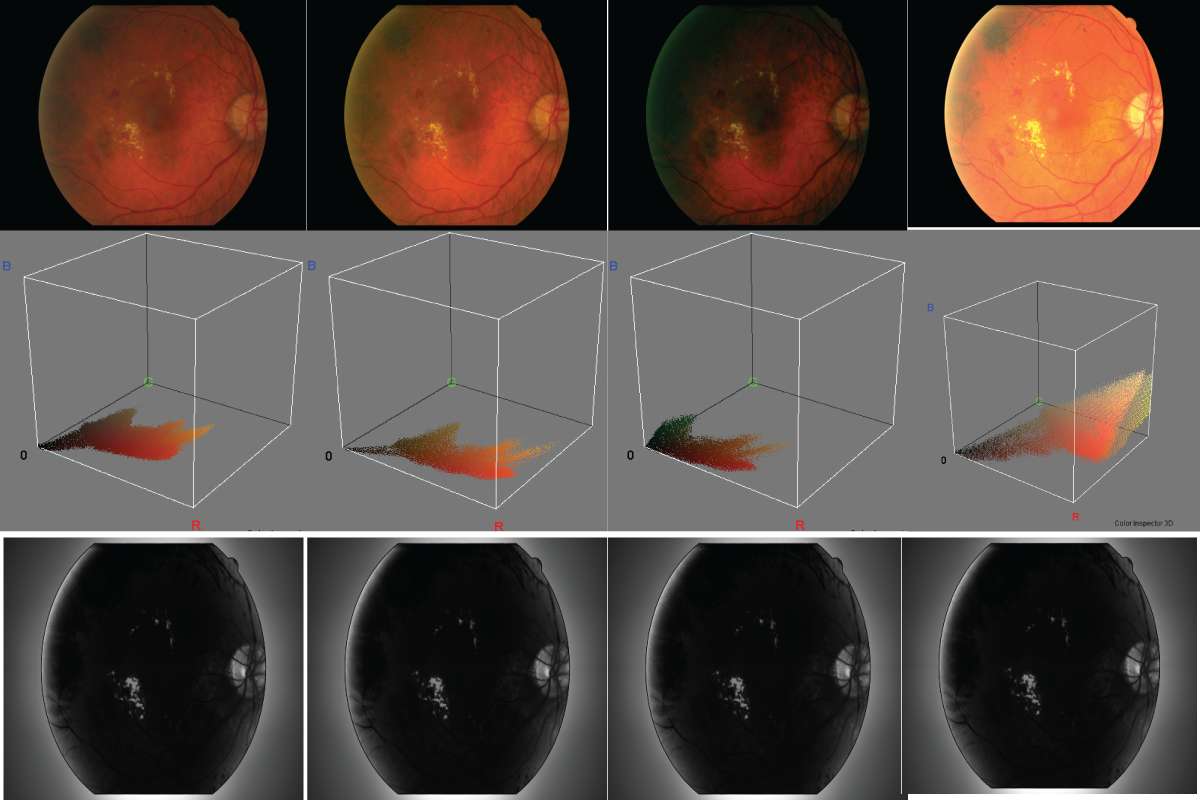}
\caption{Top-Row: Original Image1 and its different version with modified contrast and luminosity, Middle-Row: Color Distribution of Both images, Bottom-Row: Final Saliency Map}
\end{figure}

\subsection{Processing Time}
For real time application, reliable method with fast computational speed is essential. The application of the proposed method takes approximately 1.1s to a typical image of the DIARETDB1V2 database of size 1152 x 1500 in conventional computer to compute SM. And validation step take maximum of 9s. Processing time can be further reduced by implementing the method on NVIDIA GPU, since most of the computation involves can be parallelised. Table \RNum{3} shows comparisons of this method processing time per image with existing method. \\
\begin{table}[H]
  \centering
  \begin{tabular}{*{20}{c}}
\hline
Method & Resolution & time \\
\hline
Proposed Method & $1152 \times 1500$ & 10.1s \\
\hline
Abdel \cite{abdel} & $605 \times 700$ & 3.5 min \\
\hline
Foracchia \cite{forc} & $605 \times 700$ &  2min \\
\hline
Lu \cite{lu} &  $605 \times 700$ & 40s \\
\hline

\end{tabular} 
  \caption{Comparions of Processing Time Per Image}
\end{table}

\subsection{Limitations}
From experiment it has been observed that our algorithm fails in situation where optic disc deeply damaged and seems to be flat region with no difference to neighbouring regions. Lack of saliency with respect to illuminance, colour or vessels structure leads to detection failure. Figure 13 right column shows detection failure, where our algorithm end up in region having exudate and blood vessels with low detection probability. \\

The objective of this work is on the development of a computer aided system for detection of the optic disc and the macula for analysis of these features to diagnosis of various diseases. From processing time and accuracy points of view this method can be used to assist opthalmologist in detection of the OD and the macula. This method have several dvantages over existing systems. Seperation of detection and validation step inceases accuracy significantly. Also no seperate preprocessing is required for this system. In addition to that luminace and contrast invariance make this method suitable for practical purpose.

\section{CONCLUSIONS}
In this paper a novel computer aided diagnosis system is developed for retinal image analysis. For locating of the optic disc our method use the concept of saliency region and specific vasculature structure in the optic disc region. This system made use of unsupervised PLSA algorithm. We have achieved state of art accuracy for the optic disc detection in comparison to other algorithm. Detection of the macula based on clinical information of its location with respect to the optic disc and main course of the blood vessels

\addtolength{\textheight}{-12cm}   





\ifCLASSOPTIONcaptionsoff
  \newpage
\fi

\end{document}